\documentclass{article}

\usepackage[preprint, nonatbib]{nips_2018}

\usepackage{graphicx}
\usepackage[utf8]{inputenc} % allow utf-8 input
\usepackage[T1]{fontenc}    % use 8-bit T1 fonts
\usepackage{hyperref}       % hyperlinks
\usepackage{url}            % simple URL typesetting
\usepackage{booktabs}       % professional-quality tables
\usepackage{amsfonts}       % blackboard math symbols
\usepackage{nicefrac}       % compact symbols for 1/2, etc.
\usepackage{microtype}      % microtypography
\usepackage{subcaption}
\usepackage{cite}

\usepackage{amsmath}
\usepackage{graphicx}
\usepackage{color}

\newcommand{\graph}[1]{\mathcal{#1}}
\newcommand{\set}[1]{\mathcal{#1}}
\DeclareMathOperator{\red}{\texttt{red}}

\DeclareMathOperator{\mlp}{\texttt{mlp}}

\title{Conditional Graph Neural Processes: A Functional Autoencoder Approach}

\newcommand*\eqcon[1][\value{footnote}]{\footnotemark[#1]}
\author{
  Marcel Nassar\thanks{Equal contribution.}, Xin Wang\eqcon[1], Evren Tumer\\
  Artificial Intelligence Products Group, Intel Corporation \\
  \texttt{\{marcel.nassar, xin3.wang, evren.tumer\}@intel.com} \\
}

\begin{document}

\maketitle

\begin{abstract}

We introduce a novel encoder-decoder architecture to embed functional processes into latent vector spaces. 
This embedding can then be decoded to sample the encoded functions over any arbitrary domain. 
This autoencoder generalizes the recently introduced Conditional Neural Process (CNP) model of random processes. 
Our architecture employs the latest advances in graph neural networks to process irregularly sampled functions.
Thus, we refer to our model as Conditional Graph Neural Process (CGNP). 
Graph neural networks can effectively exploit ``local'' structures of the metric spaces over which the functions/processes are defined. 
The contributions of this paper are twofold: (i) a novel graph-based encoder-decoder architecture for functional and process embeddings, and (ii) a demonstration of the importance of using the structure of metric spaces for this type of representations. 

\end{abstract}
\section{Introduction}\label{sec:intro}

A key question in machine learning and information theory is how to learn to represent a data example. 
Autoencoders (AEs) are a popular form of representation learning algorithms composed of an encoder that maps the input into a code that \emph{represents} it (according to some criterion) and a decoder that interprets this code and reproduces the corresponding input~\cite{Kingma2013Auto-EncodingBayes}. The architecture of an AE is determined by the type of data it encodes: for instance, convolutional neural networks are a typical choice for image data. However, it is not clear how these constructs can be used to model more abstract data such as functions. For example, how can we learn a representation for a collection of sinusoidal functions (fundamental frequencies) or band-limited signals (Fourier) in a data-driven manner? 

While deep neural networks have achieved great success in recent years, training models that can generalize well to novel inputs requires a large amount of data.  Meta-learning techniques have been proposed to incorporate adaptation into a network so that it can accurately generalize to inputs that span domains absent in the training dataset.  

One recent approach to solving this problem attempts to combine Gaussian processes with neural networks~\cite{Garnelo2018a, Garnelo2018b}. Gaussian processes (GP) are a powerful Bayesian technique that can interpolate and extrapolate an underlying process given a set of observations.  It learns the second order statistics of the observed data points to an assumed statistical model of the function in order to generate the predictions at a given set of positions.  Though highly flexible, this method can be computationally expensive for inference.  Neural process (NP) models replace the computationally expensive portion of the GP model with a neural network that is trained on many samples taken from a distribution of functions~\cite{Garnelo2018a, Garnelo2018b}. NPs, like GPs, are able to capture the statistical properties of the sampled functions in the training set and generalize to fit points sampled from previously unobserved function realizations.

Graph convolution networks (GCN)~\cite{Defferrard2016, Kipf2016} have expanded the power of deep neural networks (DNN) into the space of data that can be described through connected graphs (e.g. social networks, knowledge graphs).  Extending DNNs to operate on graph data structures makes them potential solutions to a whole new set of problems.  Here, we expand the NP paradigm by employing GCNs and show that data can be embedded in a graph structure to improve learning of functions.

\section{Conditional Graph Neural Processes}\label{sec:gcnp}

\textbf{Functional Autoencoders using Graph Neural Networks\ }%\label{sec:func-ae}
Given a family of functions $\set{F}=\{f_\theta\}$ ($f_\theta:X\rightarrow Y$) and a dataset of different samples from each of these functions $\set{D}=\{\set{D}_k\}$, where $\set{D}_k=\{(x_i, y_i=f_\theta(x_i))\}_{i=0}^{N_\theta}$ for some parameter $\theta$, we would like to extract a representation summarizing each dataset $\set{D}_k$ by a representation $r_k$. A similar setup for non-function data was explored in \cite{Edwards2016TowardsStatistician}, where each dataset is a collection of samples and the learner's goal is to obtain a generative model for each.

\begin{figure}[t]
\captionsetup{aboveskip=0pt,belowskip=-3pt}
  \centering\hspace{-1.5em}
  \begin{subfigure}[b]{55mm}
    \includegraphics[width=\linewidth]{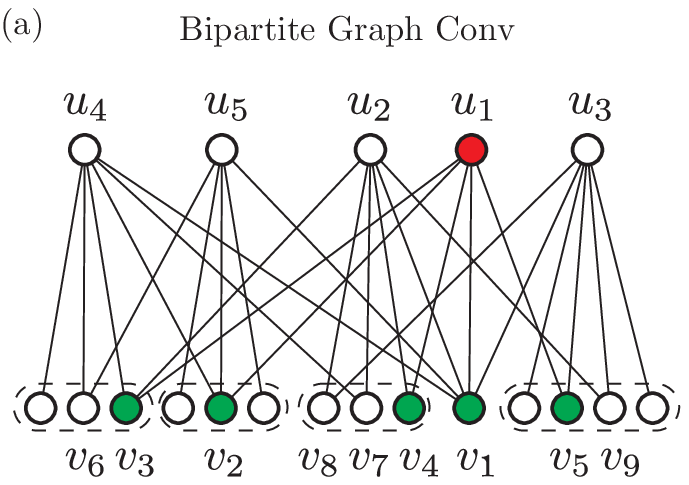} 
    \phantomsubcaption
    \label{fig:graph}
  \end{subfigure}
  \begin{subfigure}[b]{88mm}
    \includegraphics[width=\linewidth]{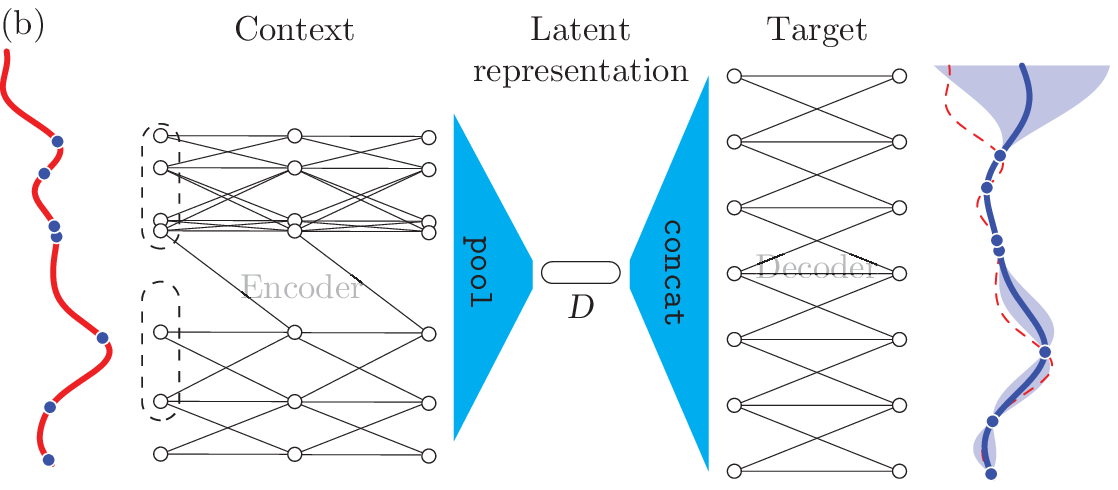} 
    \phantomsubcaption
    \label{fig:cgnp}
  \end{subfigure} 
  \vspace{-2em}
  \caption{\footnotesize
  	(\subref{fig:graph}) Bipartite graph convolution.  
  	(\subref{fig:cgnp}) Conditional Graph Neural Process.  
  	Irregularly sampled points (circles) are processed by a graph neural net encoder.  
    Unlike a CNP, graph convolution is performed over neighborhoods of the input (dashed outlines).  
    A graph pooling operation reduces the input to a $D$-dimensional latent representation, which is then "decoded" by a graph neural net decoder to generate an approximation of the function.
  }
\end{figure}

In order to process irregularly sampled data such as given in each $\set{D}_k$, we employ graph neural networks and in particular \textit{bipartite graph convolutions} introduced in \cite{Nassar2018}. Graph neural networks allow for efficient exploitation of the metric space by leveraging the induced metric as a relationship between two samples. The bipartite graph convolution operation (illustrated in Fig.\ref{fig:graph} is defined over a bipartite graph $\graph{BG}(\set{V}_i, \set{V}_o, \set{E})$
\begin{equation}\label{eq:bigconv}
g_{\graph{BG}}(v_o) = \red(\{W_{o,i}f_i| v_i \in \delta_{\graph{BG}}(v_o), f_i=s(v_i)\}), \text{    } \forall v_o \in \set{V}_o
\end{equation}
where $\red$ is a reduction operation and  $\delta_{\graph{GB}}(v_o)=\{
v_i\in\set{V}_i|(v_i, v_o)\in\set{E}\}$ is the neighborhood of the node $v_o$ in $\graph{BG}$. Our final functional autoencoder architecture is illustrated in Fig.~\ref{fig:cgnp}.

\textbf{Conditional Random Processes\ }
A general random process defines a joint probability distribution over an ordered set of random variables. A conditional neural process, on the other hand, uses the observed samples to estimate the density parameters for the targets \cite{Garnelo2018ConditionalProcesses}. This can be viewed as a more general case of the functional autoencoder model. While the functional autoencoder uses a graph neural network to output predicted values at the target points, the process version outputs a variance in addition to the mean value to characterize the uncertainty around the target point; i.e., if $g_\beta(\cdot)$ represents the function induced by the graph neural network then 
$\mu_i,\sigma_i = g_{\beta}(x_i)$ where $\mu_i,\sigma_i$ are the mean and variance around the mean prediction.

\textbf{Conditional Neural Processes\ }%\label{sec:cnp}
The CNP model described in \cite{Garnelo2018a,Garnelo2018ConditionalProcesses} uses the following to compute its representation:
\begin{align*}
r_i &= \mlp{(x_i, y_i)} \\
r &= \red{\{r_i\}_N} \\
\mu_i, \sigma_i &= \mlp{(x_t, r)}
\end{align*}

In other words, each point of the input is processed independently, then all the encoded points ($r_i$) are aggregated. The function is approximated at the target points ($x_t$) by the decoder which takes the aggregated signal concatenated with the target points as input.  Upon further inspection, a CNP can be viewed as a special case of the CGNP with an unstructured graph, where each node only connects to its self. As a result the CNP acts like a graph that is constructed using architectures that enforce invariance over transformations such as ordering  \cite{Zaheer2017DeepSets}. Here we argue that the graph structure is important for generating a better representation of the data by using the local structure that is captured by the neighborhood of the graph nodes.

\section{Experimental results}
\label{sec:res}

\textbf{Regression task\ }
We demonstrate CGNP with a simple 1-D function regression task, matching the protocol in \cite{CNP_GitHub}.  
Target functions are generated by a Gaussian process with an exponentiated quadratic kernel of length $0.4$. 

A training example $\left([\mathbf{x}_c, \mathbf{x}_t], [\mathbf{y}_c, \mathbf{y}_t]\right)$ consists of $N_c \in \{3, \cdots, 10\}$ context and $N_t \in \{2, \cdots, 10\}$ target points, independently and uniformly sampled from interval $x \in [-2, 2]$ for a specific function instance. 
The training set has $2 \times 10^5$ batches, each batch contains $64$ examples with $(N_c, N_t)$ fixed across examples in a batch.
The test set has $1 \times 10^4$ examples, each comprised of $400$ points evenly spaced in interval $[-2, 2]$, of which a randomly chosen subset of $N_c \in \{3, \cdots, 10\}$ points were used as context, and the rest $N_t = 400-N_c$ points as target.  

\textbf{Models\ }
The baseline CNP model is taken from \cite{CNP_GitHub} with the encoder and decoder composed of 3-layer and 2-layer multi-layer perceptron (MLP), respectively.  
The parameter $D$ represents the dimensionality of the latent representation.
Furthermore, we added pre-activation batch-normalization layers, which proved to improve model performance.  
Our GCNP model follows the published CNP architecture in terms of the encoder and decoder depths ($3$ and $2$) and width ($D$), but replaces the MLP networks with bipartite graph convolutional networks~\cite{Nassar2018}.  
The radius of graph connection neighborhood is set to be $\rho = 0.7$.  
For comparison we included results for a disconnected CGNP (i.e. $\rho = 0$), which reduces the CGNP to a CNP equivalent. 

Models were trained on the entire training set with Adam optimizer at learning rate $10^{-3}$, and tested for negative log-likelihood (NLL) loss and prediction mean squared error (MSE) on the test set.

\textbf{Results\ }
We list all test metrics of the three models for the regression tasks in Table~\ref{tb:res}.  
We used latent feature dimension $D = 8$ for all models.  
We did $5$ runs of CNP model, and report mean\,$\pm$\,standard deviation as a baseline range.  

A CGNP model with neighborhood radius $\rho = 0.7$ significantly improved test performance, and further setting $\rho = 0$ turned CGNP into a special case (i.e. CNP) with edgeless (i.e. disconnected) graphs, performance returned to roughly the baseline CNP level.

\begin{table}[t]
\caption{
  Test NLL and MSE of CNP and GNCP models
}
\label{tb:res}
\centering
\begin{tabular}{ l | l | l }
  \toprule
  %\multirow{2}{*}{Model} 
  \bf Model
  %& \multicolumn{2}{c|}{Gaussian Process} & \multicolumn{2}{c}{Sine wave} \\ 
%\cmidrule{2-3}
  %& \multicolumn{1}{c|}{Test NLL} & \multicolumn{1}{c|}{Test MSE}
  & \multicolumn{1}{c|}{\bf Test NLL} & \multicolumn{1}{c}{\bf Test MSE} \\
  \midrule
  CNP ($D=8$)               & $93.52 \pm 8.32$     & $0.5517 \pm 0.0120$    \\%& $27.09 \pm 13.81$     & $0.3818 \pm 0.0080$ \\
  CGNP ($D=8, \rho=0.7$)    & $56.42$     & $0.5351$     \\%& $1.06$     & $0.4075$     \\
  CGNP ($D=8, \rho=0$)      & $69.02$     & $0.5479$     \\%& $-$     & $-$  \\  
  \bottomrule
\end{tabular}  
\end{table}

\section{Discussion}

Above we have demonstrated the power of graph convolutions to incorporate local structure of context points to improve the performance of CNPs for functional fitting.  
Thus, CGNP breaks free from CNP's constraint of permutation-invariance in aggregating encoded contexts, making it able to exploit meaningful structure inherent in the context. 
As a special case, by making the graph neighborhoods used in the convolutions include only a single node, the CGNP reduces to a CNP.

This generality of CGNP over CNP is conceivably useful in modeling non-stationary data, such as those generated by Gaussian processes with time-dependent kernels.  Local structure of such functions over time can no longer be captured at a global level but will be required to interpolate the function in local neighborhoods.

\bibliographystyle{plain}
\bibliography{refs}

\end{document}